\documentclass[letterpaper]{article} % DO NOT CHANGE THIS
\usepackage{aaai2026}  % DO NOT CHANGE THIS
\usepackage{times}  % DO NOT CHANGE THIS
\usepackage{helvet}  % DO NOT CHANGE THIS
\usepackage{courier}  % DO NOT CHANGE THIS
\usepackage[hyphens]{url}  % DO NOT CHANGE THIS
\usepackage{graphicx} % DO NOT CHANGE THIS
\usepackage{natbib}  % DO NOT CHANGE THIS AND DO NOT ADD ANY OPTIONS TO IT
\usepackage{caption} % DO NOT CHANGE THIS AND DO NOT ADD ANY OPTIONS TO IT
\usepackage{algorithm}
\usepackage{algorithmic}
\usepackage{amsmath}
\usepackage{amssymb}
\usepackage{multirow}
\usepackage{booktabs}
\usepackage{newfloat}
\usepackage{listings}
\usepackage{tcolorbox}
\DeclareCaptionStyle{ruled}{labelfont=normalfont,labelsep=colon,strut=off} % DO NOT CHANGE THIS
\floatstyle{ruled}
\newfloat{listing}{tb}{lst}{}
\floatname{listing}{Listing}
\nocopyright
\title{AURA: Development and Validation of \\ an Augmented Unplanned Removal Alert System using Synthetic ICU Videos}
\author{
    Junhyuk Seo\textsuperscript{\rm 1, 2},
    Hyeyoon Moon\textsuperscript{\rm 1},
    Kyu-Hwan Jung\textsuperscript{\rm 1},
    Namkee Oh\textsuperscript{\rm 3},
    Taerim Kim\textsuperscript{\rm 1, 4, \thanks{Corresponding author.}}
}

\affiliations{
    \textsuperscript{\rm 1}Samsung Advanced Institute for Health Sciences \& Technology, Sungkyunkwan University \\
    \textsuperscript{\rm 2}Department of Nursing, Samsung Medical Center \\
    \textsuperscript{\rm 3}Department of Surgery, Samsung Medical Center, Sungkyunkwan University School of Medicine \\
    \textsuperscript{\rm 4}Department of Emergency Medicine, Samsung Medical Center, Sungkyunkwan University School of Medicine \\
    \{junhyuk.sam, moon012427, kyuhwanjung, ngnyou, taerim.j.kim\}@gmail.com
}

\begin{document}

\maketitle

\begin{abstract}

Unplanned extubation (UE)—the unintended removal of an airway tube—remains a critical patient safety concern in intensive care units (ICUs), often leading to severe complications or death. 
Real-time UE detection has been limited, largely due to the ethical and privacy challenges of obtaining annotated ICU video data. 
We propose Augmented Unplanned Removal Alert (AURA), a vision-based risk detection system developed and validated entirely on a fully synthetic video dataset. 
By leveraging text-to-video diffusion, we generated diverse and clinically realistic ICU scenarios capturing a range of patient behaviors and care contexts. 
The system applies pose estimation to identify two high-risk movement patterns: collision, defined as hand entry into spatial zones near airway tubes, and agitation, quantified by the velocity of tracked anatomical keypoints. 
Expert assessments confirmed the realism of the synthetic data, and performance evaluations showed high accuracy for collision detection and moderate performance for agitation recognition.
This work demonstrates a novel pathway for developing privacy-preserving, reproducible patient safety monitoring systems with potential for deployment in intensive care settings.
\end{abstract}

\begin{links}
    \link{Code}{https://github.com/seo-see/AURA}
    \link{Datasets}{https://doi.org/10.5281/zenodo.17577177}
\end{links}
\section{Introduction}

Mechanical ventilation via artificial airway is a lifesaving intervention for critically ill patients in intensive care units (ICUs)~\cite{chao2017multidisciplinary}. 
Paradoxically, one of the most critical safety concerns in ICUs is unplanned extubation (UE)—the unintended removal of the airway tube~\cite{christie1996unplanned}. 
A systematic review reported that approximately 6.7\% of mechanically ventilated patients experience UE, and over 80\% of these cases are self removal~\cite{li2022unplanned}. 
Such events are associated with prolonged ICU stays, increased healthcare costs, higher rates of infection, and elevated mortality~\cite{da2012unplanned}. 
In response, healthcare providers have implemented various preventive strategies, including physical restraints, standardized sedation protocols, and routine delirium screening~\cite{chang2008influence, jarachovic2011role, alaterre2023monitoring}.

Several studies have identified nurse unavailability at the bedside as the dominant factor contributing to UE, accounting for nearly 90\% of incidents~\cite{curry2008characteristics, selvan2014self}, and insufficient nurse staffing has been consistently linked to higher UE rates~\cite{penoyer2010nurse}. 
These findings indicate that continuous bedside oversight is critical for UE prevention but is difficult to sustain in real-world ICUs. 
The COVID-19 pandemic further exacerbated this challenge, as staffing shortages and strict isolation protocols hindered patient monitoring, coinciding with an increased UE incidence~\cite{berkow2020covid, taylor2021challenges}. 
In these low-resource and high-burden settings, video monitoring has been proposed as a potential solution~\cite{taylor2021challenges}.

However, conventional video monitoring relies on human observers and operates passively~\cite{cournan2016improving, abbe2021continuous}. 
Although active patient monitoring using wearable sensors and Internet of Things (IoT) devices has been explored for patient safety~\cite{chiang2019cmos, joseph2024regressive}, these methods impose additional maintenance burdens on clinical staff. 
More recently, computer vision approaches have shown promise for non-contact, sensor-free patient monitoring. 
While fall detection and risk assessment have been widely studied to improve patient safety~\cite{ gharghan2024comprehensive, gabriel2025continuous}, UE-specific applications remain limited, largely due to the ethical and privacy concerns of ICU surveillance video~\cite{chen2024spatio}.

To address these gaps, we propose Augmented Unplanned Removal Alert (AURA), a system that leverages synthetic ICU video data to detect UE risk. By using visually realistic yet privacy-preserving synthetic footage, AURA enables system development and expert evaluation without real ICU recordings—allowing refinement and failure mode analysis.
The system relies on pose estimation of generalizable anatomical cues (e.g., mouth–hand proximity) instead of hand-crafted features, enabling future extension to other critical devices such as central venous catheters. At its current stage, AURA functions as a pre-deployment tool, demonstrating how synthetic video can enable privacy-preserving, reproducible development of real-time safety monitoring systems for future clinical integration.

\begin{figure*}[ht!]
    \centering
    \includegraphics[width=\textwidth]{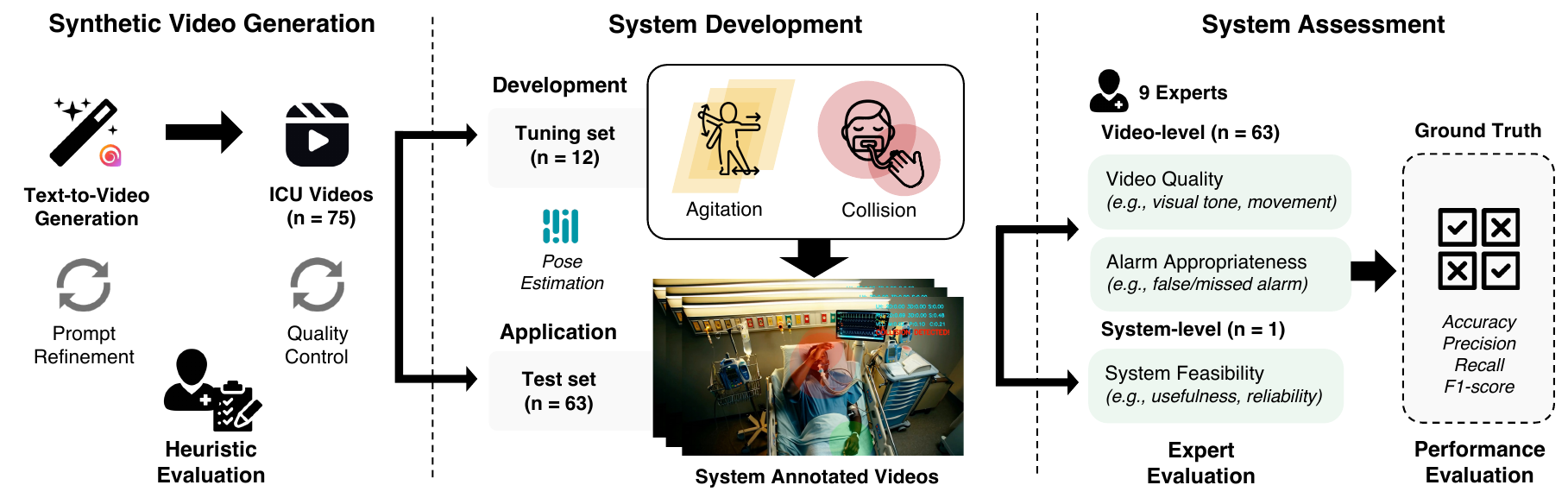}
    \caption{
Overview of the AURA development pipeline. The framework consists 
of three stages: (1) Synthetic video generation: 75 ICU videos 
generated using text-to-video model and refined through heuristic 
evaluation; (2) System development: Pose estimation-based detection 
of collision and agitation behaviors, with model tuning (n=12) and testing 
(n=63); (3) System assessment: Nine experts conducted video- and system-level evaluation, establishing ground truth for subsequent performance evaluation.}
    \label{fig:aura_workflow}
\end{figure*}

\section{Related Work}

\subsection{Data-Driven Approaches}
Several studies have integrated machine learning (ML) models into clinical workflows to mitigate the risk of unplanned tube or catheter removal. 
\citet{hur2021development} developed and validated a prediction model for UE using electronic health record (EHR) data. Random forest (RF) was reported as the best-performing model, with an area under the curve (AUC) of 0.79 and a sensitivity of 0.95. 
Similarly, \citet{zhang2023risk} proposed a support vector machine (SVM) model for central catheter unplanned removal, achieving an AUC of 0.88 and a sensitivity of 0.89.
These ML models, however, primarily function as early-warning risk scoring tools based on historical features from EHR. 
They lack real-time awareness and cannot capture the physical movements that directly precede UE, leaving a gap between predictive warning and actionable patient monitoring.

\subsection{Wearable Sensors}
Wearable sensing has also been explored for patient safety monitoring in ICUs. 
\citet{chiang2019cmos} developed a wearable infrared sensor using complementary metal–oxide–semiconductor (CMOS) technology to detect hand movements toward airway tubes and transmit proximity data wirelessly to nursing stations. 
Likewise, \citet{joseph2024regressive} applied accelerometer-based sensors for fall prevention, a conceptually similar safety task, by continuously capturing motion patterns to identify high-risk behaviors. 
While feasible, these wearable solutions require patient compliance and device maintenance, which may disrupt workflow and limiting large-scale deployment.

\subsection{Computer Vision Approaches}
Computer vision applications provide a non-contact alternative by directly observing patient behavior. 
\citet{chen2024spatio} proposed a hand-crafted spatio-temporal feature framework that combined corner detection, trajectory distance, and wavelet transforms to characterize motion patterns. Among various models, SVM model achieved an accuracy of 0.85 for binary classification (UE tendency vs. no UE tendency) and 0.62 for three-class classification (including no movement). Although this approach demonstrated competitive performance, it required labor-intensive feature engineering and relied on limited, privacy-sensitive ICU video data, restricting both scalability and reproducibility.

\subsection{Our Contributions}
In contrast to previous approaches that rely on structured EHR data, 
wearable sensors, or real ICU surveillance footage, our proposed system 
introduces a fully synthetic, vision-based framework for real-time UE 
risk detection. We leverage pose estimation on synthetic videos to 
detect high-risk behaviors without the need for physical sensors, 
hand-crafted features. This enables 
scalable, reproducible real-time safety monitoring while preserving patient privacy.

Our system is also distinct in that it incorporates structured clinical 
expert evaluations, an underexplored aspect in prior UE risk detection research.
Through expert assessments of system reliability and clinical feasibility, 
we establish a clinically grounded validation framework that enhances the interpretability and trustworthiness of the system prior to deployment.

\section{Data and Methods}
This section describes the development and validation of our synthetic 
ICU video dataset and the proposed vision-based system for UE risk detection. The process comprises three main stages: (1) Synthetic video generation, (2) System development, and (3) System assessment.
Figure~\ref{fig:aura_workflow} provides an overview of this workflow.

\subsection{Synthetic Video Generation and Dataset}
The AURA dataset comprises 75 synthetic ICU videos depicting intubated patients. Each video is 6 seconds long, with a frame rate of 25 frame per second (fps) and a resolution of 1280×720 pixels. The videos were generated using Hailuo AI’s T2V-01-Director model~\cite{hailuoai2024}. This model was selected for its ability to produce static, surveillance-like scenes, while other tested models tended to generate continuously shifting, cinematic camera movements that were unsuitable for this study’s objective.

Video generation prompts were carefully designed and validated by an experienced ICU nurse. A structured prompt template was used to systematically create diverse scenarios varying by patient gendered appearance and racial appearance, behavior (e.g., calm, agitated, lifting hands, pulling tube), and the presence of medical staff in the scene. To enhance visual diversity, we explicitly included Hispanic/Latino as a distinct appearance category. We limited our dataset to 75 videos to balance diversity with practical constraints while ensuring adequate coverage of key variations and keeping the evaluation feasible for experts.

\begin{figure}[hb]
    \centering
    \includegraphics[width=0.9\linewidth,keepaspectratio]{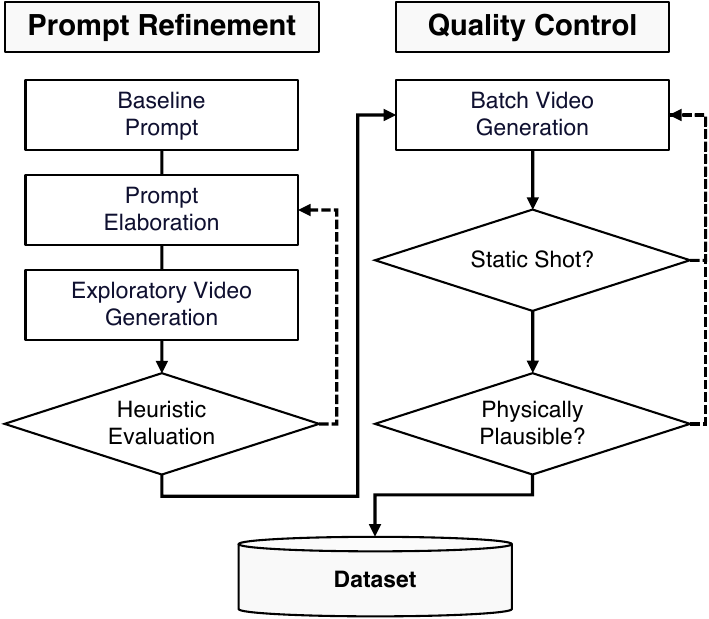}
    \caption{Overview of the AURA synthetic video dataset generation. 
    Prompts are refined via exploratory generation and heuristic expert evaluation, then used for batch video generation. 
    Outputs pass two screening stages and only accepted videos form the final dataset (solid = pass; dashed = feedback/fail). This integrated human–AI workflow ensures realism and feasibility while preserving reproducibility.}
    \label{fig:aura_visualization}
\end{figure}

The generation process involved iterative prompt refinement through heuristic evaluation, including exploratory generation and quality control, to ensure realistic and contextually accurate ICU scenes (Figure \ref{fig:aura_visualization}). 
Prompts were tuned through small-scale generations, and outputs exhibiting cinematic shots or physically implausible patient movements were excluded. This process maintained both visual fidelity and reproducibility, providing a robust foundation for subsequent video- and system-level validation.

Each video required approximately two minutes to generate, and about 20\% of the outputs failed quality control and were re-generated. Importantly, no real-world resources (such as photographs or videos) were used as inputs during generation. The complete set of synthetic videos, along with representative prompts, is publicly available on Zenodo and a summary of the final dataset is shown on Table \ref{tab:dataset_attributes}. Notably, the dataset’s gendered appearance distribution (56\% male, 44\% female) closely mirrors that reported among ICU patients worldwide~\cite{merdji2023sex}.

\begin{table}[ht]
\centering
\caption{Descriptive statistics of the full synthetic ICU video dataset ($n = 75$). All appearance attributes reflect visually generated characteristics.}
\begin{tabular}{lcc}
\hline
\textbf{Attribute} & \textbf{Count (n)} & \textbf{Proportion (\%)} \\
\hline
Gendered appearance & & \\
\quad Male & 42 & 56.0 \\
\quad Female & 33 & 44.0 \\
\hline
Racial appearance & & \\
\quad Black & 21 & 28.0 \\
\quad Asian & 20 & 26.7 \\
\quad White & 20 & 26.7 \\
\quad Hispanic/Latino & 14 & 18.7 \\
\hline
Behavior & & \\
\quad Movement & 49 & 65.3 \\
\quad Calm & 26 & 34.7 \\
\hline
Camera angle & & \\
\quad Vertical & 61 & 81.3 \\
\quad Diagonal & 7 & 9.3 \\
\quad Horizontal & 7 & 9.3 \\
\hline
Full body visibility & & \\
\quad No & 66 & 88.0 \\
\quad Yes & 9 & 12.0 \\
\hline
Medical staff & & \\
\quad None & 64 & 85.3 \\
\quad Interaction & 8 & 10.7 \\
\quad Presence & 3 & 4.0 \\
\hline
\end{tabular}
\label{tab:dataset_attributes}
\end{table}

\subsection{System Development}

The purpose of this study is to detect risk of unplanned airway removal in ICU patients. We implemented a pose estimation model to enable future extensions beyond airway monitoring, including detection of risks related to other critical devices (e.g., central venous catheters and chest tubes). All thresholds were fixed by clinical expert after optimization on a 12-video tuning set. We then validated on 63 unseen videos to avoid leakage.

Multiple studies have identified agitation as the most significant patient-related risk factor for UE~\cite{hur2021development}. Agitation, a key clinical indicator of UE risk, is formally captured by the Richmond Agitation-Sedation Scale (RASS)~\cite{sessler2002richmond}.
According to the RASS, agitation ranges from restless and anxious behavior to overtly combative actions. A very agitated state often includes direct risk behaviors such as "pulling on or removing tubes," whereas moderately agitated states manifest as anxiousness, restlessness, and frequent non-purposeful limb movements.

\begin{figure*}[t]
    \centering
    \includegraphics[width=\textwidth,keepaspectratio]{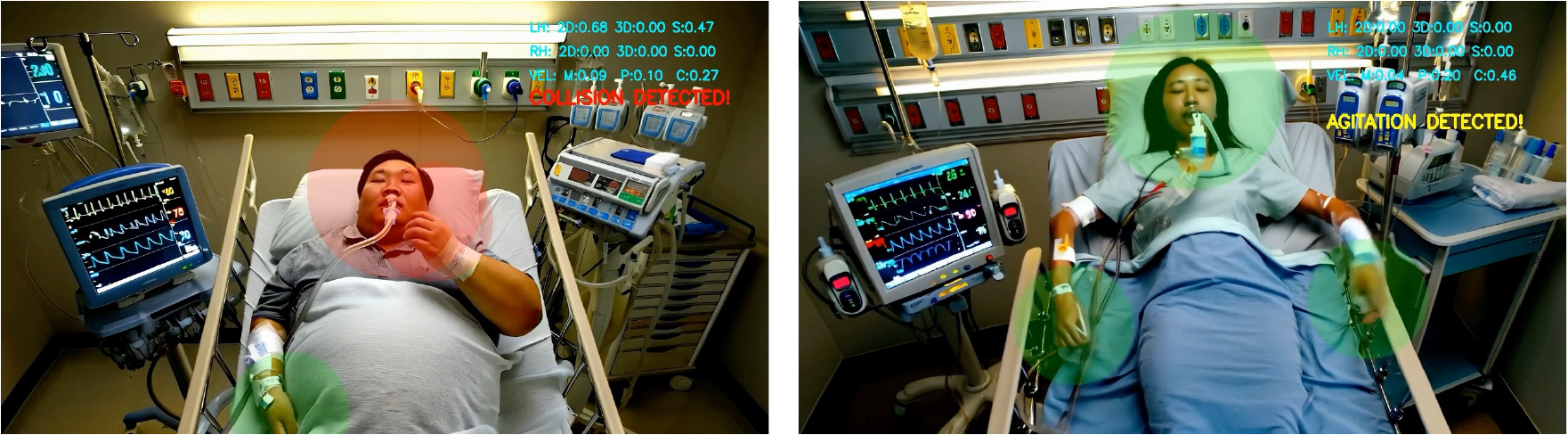}
    \caption{Examples of overlays for collision (left) and agitation (right) in synthetic intensive care unit videos.
    Colored “auras” around hands and mouth indicate risk zones (green = normal; red = collision).
    Numeric overlays show hand proximity (LH, RH), movement velocity (VEL).}
    \label{fig:aura_example}
\end{figure*}

To capture both immediate and early risks, we developed two behavior detection mechanisms: \textbf{collision}, which indicates purposeful hand movements toward the airway tube and requires immediate intervention, and \textbf{agitation}, a broader behavioral state that may predispose the patient to collision events. This distinction also reflects practical differences in detection: collision is spatially well-defined and more reliably detectable, while agitation is inherently more abstract and challenging to capture (Figure~\ref{fig:aura_example}).

The \textbf{collision} detection algorithm integrates 2D overlap analysis and 3D proximity measurement to address the limitations of 3D depth estimation from monocular cameras. While 3D coordinates provide depth information, they are often noisy and unreliable due to variability of camera angle and full body invisibility. Therefore, we complement 3D proximity with 2D overlap analysis, which yields more stable and robust measurements in the image plane.

The 2D overlap score between mouth and hand auras is calculated as:

\begin{equation*}
\text{overlap\_score} = \max\left(0, 1 - \frac{d_{2D}}{r_h + r_m}\right)
\end{equation*}

\vspace{1em}

where $d_{2D} = \sqrt{(x_h - x_m)^2 + (y_h - y_m)^2}$ is the 2D distance between hand and mouth coordinates, and $r_h$, $r_m$ are the hand and mouth aura radii respectively.

The 3D proximity score is computed as:

\begin{equation*}
\text{proximity\_score} = \max\left(0, 1 - \frac{d_{3D}}{\tau_{base}}\right)
\end{equation*}

where $d_{3D} = \sqrt{(x_h - x_m)^2 + (y_h - y_m)^2 + (z_h - z_m)^2}$ is the 3D distance and $\tau_{base}$ is the distance threshold.

The final collision score combines both metrics with weighted averaging:

\begin{equation*}
\text{score} = \alpha \cdot \text{overlap\_score} + \beta \cdot \text{proximity\_score}
\end{equation*}

\vspace{1em}

where $\alpha$ and $\beta$ denote the respective weights. A collision is triggered when the final score exceeds the threshold $\tau_{score}$. To mitigate false alarms caused by anchor instability, we apply a persistence-based mechanism that confirms a collision only if the risk state lasts longer than a minimum duration ($\tau_{duration}$). In addition, keypoints with visibility scores below $\tau_{valid}$ are excluded from the analysis to reduce noise from low-confidence detections.

For \textbf{agitation} detection, we quantify movement intensity based on the 3D velocity of each keypoint $i$, defined as $v_i$, 
where $d_i$ represents the 3D displacement between consecutive frames:
\begin{align*}
v_i &= \frac{d_i}{\Delta t}, \\[3pt]
d_i &= \sqrt{(x_i^{t} - x_i^{t-1})^2 + (y_i^{t} - y_i^{t-1})^2 + (z_i^{t} - z_i^{t-1})^2}.
\end{align*}

Only keypoints with visibility $\geq \tau_{valid}$ are included to ensure reliability in motion estimation. 
For a temporal window of $w$ frames, we derive three motion statistics:
\begin{align*}
\text{mean\_velocity} &= \frac{1}{|V_{valid}|} \sum_{v \in V_{valid}} v, \\
\text{peak\_velocity} &= \max_{v \in V_{valid}} v, \\
\text{cumulative\_velocity} &= \sum_{v \in V_{valid}} v.
\end{align*}

Agitation is detected when any of the following conditions are met:

\begin{align*}
\small{
\text{is\_agitation} =
\left\{
\begin{array}{ll}
\text{True} &\text{if } 
    (\text{mean\_velocity} > \tau_{\text{speed}}) \\
            &\text{or } (\text{peak\_velocity} > \tau_{\text{speed}}) \\
            &\text{or } (\text{cumulative\_velocity} > \tau_{\text{speed}} \cdot w) \\[3pt]
\text{False} &\text{otherwise.}
\end{array}
\right.
}
\end{align*}

\vspace{1em}

A key feature of AURA is the visualization of augmented spatial zones for collision detection. Circles surrounding the patient's hands and mouth—like an aura—appear green during normal status and change to red when a collision is detected. These overlays were purposefully designed to enhance clinical interpretability and situational awareness, while they do not affect the underlying detection logic.

All system parameters were optimized through iterative clinical expert evaluation with tuning set of 12 videos. The expert reviewed system performance across various scenarios and adjusted parameters to achieve optimal balance between detection sensitivity and false alarm reduction (Table \ref{tab:parameters}).

\begin{table}[h]
\centering
\caption{System Parameters Calibrated During Prototype Development}
\begin{tabular}{lll}
\hline
\textbf{Parameter} & \textbf{Value} & \textbf{Description}\\
\hline
$\tau_{base}$ & 0.3 & 3D distance threshold \\
$r_m$ & 150 pixels & Mouth Aura radius \\
$r_h$ & 100 pixels & Hand Aura radius \\
$\alpha$ & 0.7 & 2D overlap weight \\
$\beta$ & 0.3 & 3D proximity weight \\
$\tau_{score}$ & 0.3 & Final score threshold \\
$\tau_{duration}$ & 0.3s & Persistence duration \\
$\tau_{speed}$ & 0.18 & Agitation speed threshold \\
$\tau_{valid}$ & 0.7 & Keypoint visibility threshold \\
$w$ & 5 frames & Analysis window size \\
\hline
\end{tabular}
\label{tab:parameters}
\end{table}

After prototype development, the AURA system was applied to the test set of 63 unseen videos. Each video was processed through the optimized detection pipeline, generating collision and agitation alerts based on clinically calibrated thresholds. As the primary objective was feasibility validation using pre-recorded data, real-time streaming integration was not implemented at this stage.

\begin{table*}[ht]
\centering
\caption{Evaluation framework for synthetic ICU video dataset (5-point Likert, with range)}
\begin{tabular}{llcc}
\hline
\textbf{Domain} & \textbf{Definition} & \textbf{Mean ± SD} & \textbf{Range} \\
\hline
Visual tone & Visual tone (brightness/contrast) of the video is similar to clinical setting & 4.4 ± 0.1 & 4.0–4.7 \\
Background setting & Background of the video is similar to clinical setting & 4.3 ± 0.2 & 3.3–4.6 \\
Patient behavior & Behavior of the patient in the video seems natural & 4.6 ± 0.2 & 3.9–4.8 \\
Clinical plausibility & The event presented in the video is plausible in the clinical context & 4.6 ± 0.2 & 4.2–4.9 \\
\hline
\end{tabular}
\vspace{0.5em}
\label{tab:dataset_evaluation}
\end{table*}

Pose estimation was performed using MediaPipe, selected for its cross-platform efficiency and low-latency stream processing capabilities~\cite{lugaresi2019mediapipeframeworkbuildingperception}. All experiments were conducted with Python 3.9 and OpenCV on macOS (Apple M3 Max, 128GB RAM), without GPU acceleration. The source code is publicly available on our GitHub repository.

\subsection{System Assessment}

We designed a single integrated validation framework that simultaneously assessed the realism of the synthetic ICU videos and the performance of the AURA system. A structured web-based questionnaire, co-developed by clinical and informatics experts, holistically evaluated three domains:
(1) the quality of the synthetic videos,
(2) the clinical appropriateness of system-generated alarms, and
(3) the feasibility of integrating the system into real-world ICU workflows.

Nine ICU nurses independently evaluated 63 annotated videos.
The evaluators were evenly distributed across three experience levels (early-career: 1–3 years, mid-career: 3–5 years, and senior-level: over 5 years) to ensure diverse clinical perspectives. For each video, they rated four dimensions of visual realism (visual tone, background setting, patient behavior, and clinical plausibility) as well as alarm appropriateness using a 5-point Likert scale (1 = strongly disagree, 5 = strongly agree). To assess the reliability of these expert ratings, intraclass correlation coefficients (ICCs) were calculated for alarm appropriateness scores.

When an alarm was deemed clinically inappropriate, evaluators selected one of four predefined categories: false alarm, missed alarm, premature alarm, or delayed alarm. These annotations provided both qualitative insights and consensus-based reference labels for computing performance metrics (accuracy, precision, recall, and F1-score). 95\% confidence intervals (CI) were estimated using bootstrap resampling. Notably, alarms classified as premature or delayed were still counted as system-triggered alarms during evaluation, reflecting a conservative labeling strategy that prioritizes sensitivity over exact temporal alignment.

After video-level evaluation, system-level feasibility was also assessed with the same Likert scale across three criteria (clinical usefulness, reliability, and recommendation to others).

\subsection{Robustness Analysis}
Two complementary analyses were conducted to evaluate system robustness with respect to both spatial scale and parameter perturbation.
\paragraph{Scale Robustness.}
As the system initially relied on a fixed pixel aura, variations in video size, camera zoom, viewing angle, and patient body size could potentially influence model performance.  
To examine scale robustness, we tested a relative pixel aura by normalizing all distance measurements to each subject’s head and hand dimensions.  
The normalized body sizes were computed as follows:

{\small
\begin{equation*}
\begin{split}
\text{head\_size} = \max(&\mathrm{dist}(L_{\text{ear}}, R_{\text{ear}}), \\
                         &2\times \mathrm{dist}(L_{\text{eyebrow}}, L_{\text{mouth}}), \\
                         &2\times \mathrm{dist}(R_{\text{eyebrow}}, R_{\text{mouth}}))
\end{split}
\end{equation*}
}

{\small
\begin{equation*}
\begin{split}
\text{hand\_size} = 
\max(&\max(\mathrm{dist}(L_{\text{pinky}}, L_{\text{thumb}}),
           \mathrm{dist}(L_{\text{wrist}}, L_{\text{index}})), \\
     &\max(\mathrm{dist}(R_{\text{pinky}}, R_{\text{thumb}}),
           \mathrm{dist}(R_{\text{wrist}}, R_{\text{index}})))
\end{split}
\end{equation*}
}

\vspace{1em}

Let $\lambda$ denote the scaling coefficient for the aura radii.  
The final aura radii were computed as:
\begin{equation*}
\begin{aligned}
\text{head\_aura\_radius} &= \lambda \cdot \text{head\_size}, \\
\text{hand\_aura\_radius} &= \lambda \cdot \text{hand\_size}.
\end{aligned}
\end{equation*}

In our implementation, $\lambda$ was empirically set to 2.0 to balance 
sensitivity and stability in the tuning set. Additionally, we tested 
relative mode for down-scaled video (from 1280×720 to 854×480) to 
demonstrate the system robustness in low resolution video.

\paragraph{Parameter Robustness.}
We conducted a concise, 3-fold cross-validation–style sensitivity analysis. 
In each fold, 21 videos were used as a tuning set and 42 as a held-out validation set. 
Key hyperparameters were perturbed within $\pm10\%$ of their expert-defined baselines, 
and a $3^3$ grid search (27 configurations) was performed over the agitation speed threshold 
($\tau_{\text{speed}}$), visibility threshold ($\tau_{\text{valid}}$), 
and radius scaling factor ($s_r$), which scales the fixed pixel radius of the aura regions 
as $r_{\text{final}} = s_r \times r_{\text{base}}$.

The selection objective on the tuning set was the combined F1 score 
(mean of collision and agitation-F1), which was then evaluated on the held-out set. This procedure was repeated over three non-overlapping validation splits.

\section{Results}

\subsection{Expert Evaluation}
In the video quality assessment, the synthetic ICU videos were perceived as highly realistic and clinically valid. All four evaluation domains (visual tone, background setting, patient behavior, and clinical plausibility) received mean Likert scores $\geq 4.3$ (range: 3.3–4.9), with patient behavior and clinical plausibility showing the highest ratings at 4.6 (Table~\ref{tab:dataset_evaluation}).

For alarm appropriateness, mean video-wise ratings with standard deviation (SD) indicated high appropriateness for collision alarms (4.3~$\pm$~0.6) and moderate appropriateness for agitation alarms (3.8~$\pm$~0.6).
Inter-rater reliability was excellent for both alarm types, with ICC(3,k) values confirming consistent expert agreement (Table~\ref{tab:icc_results}).

\begin{table}[hb]
\centering
\caption{Inter-rater reliability for alarm appropriateness evaluation by nine experts, calculated using the intraclass correlation coefficient ICC(3,k). Degrees of freedom for the F-statistic: df\textsubscript{1} = 8, df\textsubscript{2} = 496.}
\begin{tabular}{lcccc}
\hline
\textbf{Domain} & \textbf{ICC(3,k)} & \textbf{95\% CI} & \textbf{F} & \textbf{p-value} \\
\hline
Agitation & 0.95 & [0.89, 0.99] & 19.62 & $<$0.001 \\
Collision & 0.95 & [0.89, 0.99] & 19.42 & $<$0.001 \\
\hline
\end{tabular}
\label{tab:icc_results}
\end{table}

System-level feasibility scores showed medium-to-high acceptance, with mean ratings of 3.9~$\pm$~0.9 for clinical usefulness, 3.2~$\pm$~1.0 for reliability,  
and 3.9~$\pm$~0.8 for recommendation to others.

\subsection{Performance Evaluation}
The collision detection module achieved near-perfect performance, with F1-score of 0.98 and recall of 1.00.  
In contrast, the agitation detection module demonstrated moderate performance (F1-score: 0.78), reflecting the complexity of capturing agitation behaviors (Table ~\ref{tab:aura_performance}).

\begin{table}[hb]
\centering
\caption{Performance metrics of AURA with 95\% confidence intervals.}
\begin{tabular}{lcc}
\hline
\textbf{Metric} & \textbf{Collision Detection} & \textbf{Agitation Detection} \\
\hline
Accuracy  & 0.98 [0.93, 1.00] & 0.86 [0.76, 0.95] \\
Precision & 0.96 [0.81, 0.99] & 0.89 [0.74, 1.00] \\
Recall    & 1.00 [0.86, 1.00] & 0.70 [0.51, 0.88] \\
F1-score  & 0.98 [0.93, 1.00] & 0.78 [0.62, 1.00] \\
\hline
\end{tabular}
\label{tab:aura_performance}
\end{table}

\subsection{Robustness Analysis}
The relative aura radii achieved performance comparable to the fixed-radius baseline, 
with similar F1 and recall scores for both behaviors 
(collision: F1 = 0.94, recall = 1.00; agitation: F1 = 0.78, recall = 0.70). When tested on down-scaled videos (854×480), collision detection remained highly robust (F1 = 0.92, recall = 1.00) while agitation detection showed moderate degradation (F1 = 0.65).

Across the three validation folds, 
collision detection maintained stable performance (F1 = 0.94--1.00, recall = 1.00), 
whereas agitation detection showed moderate variability (F1 = 0.69--0.83, recall = 0.60--0.75). 
The best-performing configuration was consistent across folds, 
with $\tau_{\text{speed}} = 0.18$, $\tau_{\text{valid}} = 0.63$, 
and $s_r = 0.9$. 
The overall cross-fold deviation in combined F1 was within 0.05, 
demonstrating stable performance under $\pm10\%$ parameter perturbations.

\section{Discussion}

The proposed system demonstrates the feasibility and ethical viability of developing a real-time early warning system for UE using only synthetic videos. By integrating pose estimation with realistic synthetic data, AURA bridges the gap between traditional risk-scoring models and continuous behavioral monitoring. This represents a practical and ethical alternative to real ICU video collection, overcoming privacy and logistical barriers in high-acuity settings.

Synthetic ICU videos proved to be a viable resource for both system development and expert validation. High evaluation scores (all $\geq 4.3$) across realism domains indicate that synthetic videos can replicate key visual cues necessary for risk assessment. These results validate the use of synthetic datasets for scalable and privacy-preserving development of vision-based patient monitoring tools.

The reliability of alarm appropriateness assessments, which served as the reference standard for performance evaluation, was supported by strong inter-rater agreement. The ICC(3,k) exceeded 0.95 for both collision and agitation ratings, demonstrating the robustness and consistency of the evaluation framework. In addition to validating the system’s performance, this work contributes a novel expert-annotated dataset. Given the inherently ambiguous and context-dependent nature of agitation, these annotations capture subtle clinical judgments that cannot be fully represented through rule-based definitions. This resource provides a valuable foundation for future research on behavioral monitoring and UE risk stratification.

Subsequent performance evaluation revealed a clear distinction in system accuracy across behavior types. Collision detection achieved near-perfect results (F1: 0.98; Recall: 1.00), benefiting from its spatially well-defined characteristics and the high contrast between normal and high-risk hand trajectories. In contrast, agitation detection yielded moderate performance (F1: 0.78; Recall: 0.70), reflecting the broader variability and subtlety of agitation-related movements. Future work should focus on improving quantification and categorization of agitation using advanced temporal modeling and expanded behavior taxonomies.

False alarms were primarily attributed to two sources: anchor instability and staff interference. Anchor jumps (momentary misplacement of pose keypoints near critical zones) occasionally triggered false collision alerts. Similarly, interaction between patient and medical staff in the frame led to misidentified limb movements. These issues point to limitations of the current pose estimation backend (MediaPipe), which lacks temporal smoothing and multi-person detection. Incorporating temporal filtering and gating mechanisms, as proposed in recent work~\cite{chen2024spatio}, may reduce false positives and improve clinical reliability.

Beyond quantitative performance, expert feedback also revealed the system’s practical limitations, particularly regarding alert precision. The overall clinical reliability was rated moderate (3.2/5), reflecting the high standards of accuracy expected by ICU professionals.
These findings emphasize the importance of aligning algorithmic detection thresholds with clinical expectations. Accordingly, this study introduces one of the first clinician-centered evaluation frameworks for UE detection, providing a replicable foundation for validating safety-critical monitoring systems.

Despite these practical limitations, AURA demonstrated stable detection performance when tested with relative aura radii, under $\pm10\%$ parameter perturbations, and across varying video resolutions. Collision detection remained highly accurate (F1 = 0.92) even at reduced resolution (854$\times$480, 67\% of original), indicating strong robustness to video quality degradation. Although agitation detection exhibited a moderate performance decline at lower resolutions, this is unlikely to pose significant constraints given the widespread adoption of high-definition cameras (1280$\times$720) in modern ICU surveillance systems. These findings demonstrate that the system's underlying detection logic is stable to moderate variations in both algorithmic parameters and input quality, supporting potential deployment in real-world ICUs.

Although this study evaluated system outputs on pre-recorded video, the same detection logic can be deployed in real time by subscribing to existing Closed-Circuit Television (CCTV) streams, requiring no new sensors or camera replacements. Deploying AURA in real-world ICUs would enable continuous UE risk monitoring, bringing this approach closer to clinical deployment.

To support such deployment, future iterations of AURA should incorporate adaptive thresholding, temporal stabilization, and robust multi-person tracking.
Together with real-time evaluation in actual ICU workflows, these enhancements will be essential for ensuring safe, trustworthy, and clinically reliable vision assistance in critical care settings.

\subsection{Limitations}

Our fully synthetic dataset may lack rare postures, extreme agitation, or uncommon device configurations, limiting generalizability to complex ICU scenarios. Visible phenotype categories serve solely as visual proxies and do not capture the full diversity of real-world patients.

Despite high scores for realism and plausibility, the synthetic videos may differ from real ICU footage in lighting, saturation, or camera perspective. To improve robustness, future work should incorporate diverse visual augmentations. In addition, overlays (“aura” visualizations) were visible during expert evaluations and may have influenced perceptions of video quality and event plausibility. We retained them to reduce evaluators’ cognitive load during rapid assessments, but will conduct overlay-free evaluations in future studies to better estimate potential bias.

\section{Conclusion}

We introduced AURA, a pose-estimation-based early warning system for 
UE, developed and validated entirely on a fully synthetic ICU video dataset. 
Without relying on real ICU footage or patient-worn sensors, 
AURA demonstrates the feasibility of scalable, privacy-preserving 
behavioral monitoring in critical care.

The synthetic videos achieved high realism across all evaluation domains 
(mean Likert $\geq 4.3$), and nine ICU nurses reached excellent agreement 
(ICC $>$ 0.95) when assessing system-generated alarms. 
Based on this rigorously validated ground truth, AURA achieved near-perfect collision detection (F1 = 0.98) and respectable agitation recognition (F1 = 0.78). Subsequent robustness analysis confirmed stable performance of the system with moderate variations in parameters and relative mode settings.

Our findings establish synthetic video as a valuable and viable resource 
for developing safety-critical AI in privacy-sensitive healthcare settings. 
Beyond UE monitoring, AURA’s reproducible pipeline 
provides a blueprint for ethical AI development that can be extended to 
other critical devices and patient safety applications. 
This work demonstrates that privacy-preserving approaches can deliver 
clinically meaningful, deployable monitoring solutions for intensive 
care environments worldwide.

\section{Ethics Statement}

This study was reviewed and approved by the Institutional Review Board (IRB) of Samsung Medical Center (IRB No. SMC 2025-06-135-001) and conducted in accordance with all relevant ethical standards and institutional regulations.

No real patient data or footage were used in the synthetic video generation process. 
All participating clinical experts provided written informed consent, which covered the study purpose, the collection and analysis of their evaluation responses, and the potential public release of de-identified, aggregated data (e.g., Likert-scale scores) for secondary academic use.

\section{Acknowledgments}
This research was supported by a grant of the Korea Health Technology R\&D Project
 through the Korea Health Industry Development Institute (KHIDI), funded by the Ministry of Health \&
 Welfare, Republic of Korea (grant number : RS-2025-02223382).

\bibliography{aaai2026}

\appendix
\onecolumn

\section{Appendix}
\vspace{0.5cm}

\begin{center}
\begin{tcolorbox}[
    width=0.85\linewidth,
    colback=gray!5,
    colframe=gray!40,
    fonttitle=\bfseries,
    title={Prompt Template for Video Generation}]

\begingroup
\hyphenpenalty=10000
\exhyphenpenalty=10000
\sloppy

[Background]

The room contains all characteristic ICU features: a wall-mounted medical gas panel with oxygen, compressed air, and suction ports; a multi-parameter bedside monitor showing the patient's ECG, heart rate, respiratory rate, SpO2, and blood pressure; an advanced ventilator machine with detailed circuit connections leading to the patient's endotracheal tube; multiple infusion and syringe pumps delivering medications and fluids; and a mobile equipment cart. The \texttt{$<$Racial appearance$>$} \texttt{$<$Gendered appearance$>$} patient is sedated in a fully motorized ICU bed with raised side rails. They are intubated with an endotracheal tube secured with medical tape, connected to the ventilator. Their arms have IV lines, and one wrist has an ID band. The lighting is dim, with clinical overhead lights casting a focused glow. The environment includes soft beeping sounds from the monitor and ventilator, typical alarm indicators, and clear visual signs of a critical care setting.

\vspace{0.1cm}

\textit{* Racial appearance: e.g., White, Asian, Black, Hispanic/Latino}

\textit{* Gendered appearance: e.g., male, female}

\vspace{0.3cm}

[View]

Fixed overhead \texttt{$<$Camera angle$>$} view from above the bed: A static, wide-angle overhead shot positioned directly above the foot end of the bed, looking downward toward the patient and capturing the entire bed, patient, and surrounding equipment within the frame (not angled from the head side).

\vspace{0.1cm}

\textit{* Camera angle: e.g., horizontal, vertical}

\vspace{0.3cm}

[Subject movement]

\texttt{$<$Behavior$>$}

\vspace{0.1cm}

\textit{* Behavior: e.g.,}

\textit{1) In the scene, the patient slowly raises one hand to medium height, deliberately reaching toward their mouth and the endotracheal tube (E-tube), attempting to grasp or pull it.}

\textit{2) The scene shows 2 stages of patient movement: 1) no move and calm, 2) slow, restless body movement}

\textit{3) In the scene, the patient remains calm without movement.}

\vspace{0.3cm}

[Scene constraints]

\texttt{$<$Medical staff$>$}

\vspace{0.1cm}

\textit{* Medical staff: e.g.,}

\textit{1) The video must include only the patient as the sole subject; no other people (such as medical staff, visitors, or additional patients) appear in the frame or background.}

\textit{2) A medical staff member briefly interacts with the patient.}

\vspace{0.3cm}

\textbf{[Static shot]} \textit{-- Static-shot option enabled.}
\endgroup
\end{tcolorbox}
\begin{figure}[hb]
    \centering
    \captionsetup{width=0.85\linewidth}
    \caption{
Prompt template for synthetic ICU video generation. 
Video generation was performed using an expert-designed prompt template, which was iteratively refined through sample generation and heuristic evaluation by an experienced ICU nurse.
}
    \label{fig:prompt_template}
\end{figure}
\end{center}

\newpage

\begin{table*}[t]
\small
\renewcommand{\arraystretch}{1.3}
\hyphenpenalty=10000
\exhyphenpenalty=10000
\centering
\captionsetup{width=\linewidth}
\caption{Video-level expert evaluation questionnaire for synthetic video and system assessment}
\begin{tabular}{p{3.6cm} p{9.4cm} c c c c c}
\toprule
\textbf{Item} & \textbf{Description} & \textbf{5} & \textbf{4} & \textbf{3} & \textbf{2} & \textbf{1} \\
\midrule
\multicolumn{7}{l}{\textbf{Video Realism \& Fidelity}} \\
\midrule
Visual tone 
& The visual tone (including brightness and contrast) of the video is similar to clinical setting.
& $\square$ & $\square$ & $\square$ & $\square$ & $\square$ \\
Background setting 
& The background of the video is similar to clinical setting.
& $\square$ & $\square$ & $\square$ & $\square$ & $\square$ \\
Patient behavior 
& The behavior of the patient in the video seems natural.
& $\square$ & $\square$ & $\square$ & $\square$ & $\square$ \\
Clinical plausibility 
& The event presented in the video is plausible in the clinical context.
& $\square$ & $\square$ & $\square$ & $\square$ & $\square$ \\
\midrule
\multicolumn{7}{l}{\textbf{Alarm Appropriateness}} \\
\midrule
Collision
& The collision alarm activated appropriately for the situation.
& $\square$ & $\square$ & $\square$ & $\square$ & $\square$ \\
& \textit{e.g., the alarm did not activate when the patient's hand did not approach the tube area} & & & & & \\
& If inappropriate, reason:  & & & & & \\
& \textcircled{1} Too fast \textcircled{2} Too slow \textcircled{3} Alarm activated unnecessarily \textcircled{4} Alarm failed to activate when needed & & & & & \\
Agitation
& The agitation alarm activated appropriately for the situation.
& $\square$ & $\square$ & $\square$ & $\square$ & $\square$ \\
& \textit{e.g., the alarm did not activate when the patient did not exhibit frequent movements} & & & & & \\
& If inappropriate, reason:  & & & & & \\
& \textcircled{1} Too fast \textcircled{2} Too slow \textcircled{3} Alarm activated unnecessarily \textcircled{4} Alarm failed to activate when needed & & & & & \\
\midrule
\multicolumn{7}{l}{\textbf{Suggestion}} \\
\midrule
Suggestion & \multicolumn{6}{l}{Please leave suggestions for video and system improvements.} \\
\multicolumn{7}{l}{} \\
\bottomrule
\end{tabular}
\vspace{2mm}
\makebox[\textwidth][r]{%
\footnotesize
\textit{Note:} 5 = Strongly agree, 4 = Agree, 3 = Neutral, 2 = Disagree, 1 = Strongly disagree
}
\label{tab:video_level_questionnaire}
\renewcommand{\arraystretch}{1.0}
\end{table*}

\begin{table*}[t]
\small
\renewcommand{\arraystretch}{1.3}
\hyphenpenalty=10000
\exhyphenpenalty=10000
\centering
\captionsetup{width=0.95\linewidth}
\caption{System-level expert evaluation questionnaire}
\begin{tabular}{p{3.6cm} p{9.4cm} c c c c c}
\toprule
\textbf{Item} & \textbf{Description} & \textbf{5} & \textbf{4} & \textbf{3} & \textbf{2} & \textbf{1} \\
\midrule
\multicolumn{7}{l}{\textbf{System Feasibility}} \\
\midrule
Usefulness
& This alarm system is clinically useful.
& $\square$ & $\square$ & $\square$ & $\square$ & $\square$ \\
Reliability
& This system can be trusted in clinical settings.
& $\square$ & $\square$ & $\square$ & $\square$ & $\square$ \\
Recommendation
& I would recommend this system to other healthcare professionals.
& $\square$ & $\square$ & $\square$ & $\square$ & $\square$ \\
\midrule
\multicolumn{7}{l}{\textbf{Suggestion}} \\
\midrule
Suggestion & \multicolumn{6}{l}{Please leave suggestions for video and system improvements.} \\
\multicolumn{7}{l}{} \\
\bottomrule
\end{tabular}
\vspace{2mm}
\makebox[\textwidth][r]{%
\footnotesize
\textit{Note:} 5 = Strongly agree, 4 = Agree, 3 = Neutral, 2 = Disagree, 1 = Strongly disagree
}
\label{tab:system_level_questionnaire}
\renewcommand{\arraystretch}{1.0}
\end{table*}

\newpage

\begin{figure}[ht]
    \centering
    \includegraphics[width=0.85\textwidth,keepaspectratio]{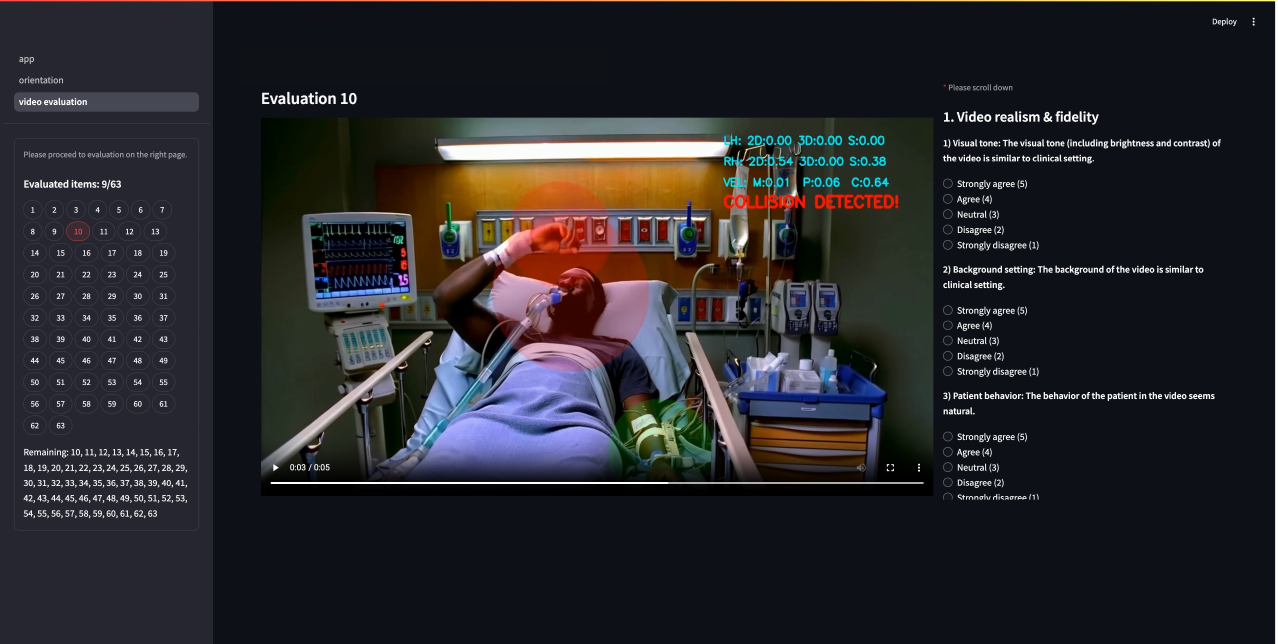}
    \captionsetup{width=0.85\linewidth}
    \caption{
Web-based evaluation interface used by ICU nurses.
The interface displays each synthetic ICU video with pose-estimated overlays (center), 
provides structured rating criteria for video quality and alarm appropriateness (right), 
and allows navigation across all 63 evaluation items (left).
Before entering the evaluation page, experts received a brief orientation explaining the system and rating procedure. 
After the video-level assessment, a system-level evaluation page was presented. 
This interface was used to collect expert assessments for validating video quality and overall system performance.
}
    \label{fig:aura_interface}
\end{figure}

\begin{table*}[ht]
\centering
\captionsetup{width=0.9\linewidth}
\caption{Demographic characteristics of ICU nurse evaluators (n = 9)}
\begin{tabular}{p{4cm} p{3cm} c}
\toprule
\textbf{Variable} & \textbf{Category} & \textbf{n (\%)} \\
\midrule

\multirow{2}{*}{Age} 
& 20s & 6 (66.7\%) \\
& 30s & 3 (33.3\%) \\
\midrule

\multirow{2}{*}{Gender} 
& Female & 6 (66.7\%) \\
& Male & 3 (33.3\%) \\
\midrule

\multirow{3}{*}{Clinical experience} 
& 1--3 years & 3 (33.3\%) \\
& 3--5 years & 3 (33.3\%) \\
& $\ge$ 5 years & 3 (33.3\%) \\
\midrule

\multirow{5}{*}{Exposure to synthetic videos}
& Never & 1 (11.1\%) \\
& Occasionally & 3 (33.3\%) \\
& Sometimes & 4 (44.4\%) \\
& Often & 0 (0.0\%) \\
& Very often & 1 (11.1\%) \\
\bottomrule

\end{tabular}
\label{tab:evaluator_demographics}
\end{table*}

\newpage
\clearpage

\begin{table}[t]
\centering
\caption{Summary of ground truth labels for collision and agitation behaviors (n = 63).}
\begin{tabular}{lcc}
\toprule
Behavior & Positive (1) & Negative (0)\\
\midrule
Collision & 24 (38.1\%) & 39 (61.9\%) \\
Agitation & 23 (36.5\%) & 40 (63.5\%) \\
\bottomrule
\end{tabular}
\end{table}

\begin{table*}[ht]
\centering
\scriptsize
\captionsetup{width=0.85\linewidth}
\caption{Robustness evaluation across scale, resolution, and parameter settings. 
Metrics include Accuracy (Acc), Precision (Prec), Recall (Rec), and F1 for both Collision and Agitation detection.}
\label{tab:robustness}
\begin{tabular}{lllcccccccc}
\toprule
\multirow{2}{*}{Domain} & \multirow{2}{*}{Mode} & \multirow{2}{*}{Setting} &
\multicolumn{4}{c}{Collision} & \multicolumn{4}{c}{Agitation} \\
\cmidrule(lr){4-7}\cmidrule(lr){8-11}
& & & Acc & Prec & Rec & F1 & Acc & Prec & Rec & F1 \\
\midrule
Baseline  & Original                & fixed aura scale; 1280$\times$720 pixels
          & 0.98 & 0.96 & 1.00 & 0.98 & 0.86 & 0.89 & 0.70 & 0.78 \\
\midrule
Aura scale & Relative radius scaling & $\lambda = 1$
           & 0.83 & 1.00 & 0.54 & 0.70 & 0.86 & 0.89 & 0.70 & 0.78 \\
Aura scale & Relative radius scaling & $\lambda = 1.5$
           & 0.92 & 0.95 & 0.83 & 0.89 & 0.86 & 0.89 & 0.70 & 0.78 \\
Aura scale & Relative radius scaling & $\lambda = 2$
           & 0.95 & 0.89 & 1.00 & 0.94 & 0.86 & 0.89 & 0.70 & 0.78 \\
Aura scale & Relative radius scaling & $\lambda = 2.5$
           & 0.94 & 0.86 & 1.00 & 0.92 & 0.86 & 0.89 & 0.70 & 0.78 \\
\midrule
Video scale & Resolution downscaling & $\lambda = 2$, 854$\times$480 pixels
            & 0.94 & 0.89 & 0.96 & 0.92 & 0.75 & 0.65 & 0.65 & 0.65 \\
\midrule
Parameter & 3-fold parameter search & Original (Fold 1)
          & 0.95 & 0.86 & 1.00 & 0.92 & 0.91 & 0.88 & 0.88 & 0.88 \\
Parameter & 3-fold parameter search & Original (Fold 2)
          & 0.95 & 0.91 & 1.00 & 0.95 & 0.76 & 0.71 & 0.63 & 0.67 \\
Parameter & 3-fold parameter search & Original (Fold 3)
          & 1.00 & 1.00 & 1.00 & 1.00 & 0.76 & 0.67 & 0.57 & 0.62 \\
Parameter & 3-fold parameter search & Relative $\lambda = 2$ (Fold 1)
          & 0.95 & 1.00 & 0.89 & 0.94 & 0.81 & 0.82 & 0.60 & 0.69 \\
Parameter & 3-fold parameter search & Relative $\lambda = 2$ (Fold 2)
          & 0.93 & 0.82 & 1.00 & 0.90 & 0.88 & 0.92 & 0.73 & 0.82 \\
Parameter & 3-fold parameter search & Relative $\lambda = 2$ (Fold 3)
          & 0.93 & 0.84 & 1.00 & 0.91 & 0.88 & 0.92 & 0.75 & 0.83 \\
\bottomrule
\end{tabular}
\end{table*}

\end{document}